\documentclass[conference]{IEEEtran}
\IEEEoverridecommandlockouts
\usepackage{cite}
\usepackage{amsmath,amssymb,amsfonts}
\usepackage{algorithm}
\usepackage{algpseudocode}
\usepackage{array}
\usepackage{arydshln}
\usepackage{graphicx}
\usepackage{textcomp}
\usepackage{subcaption}
\usepackage{xcolor}
\usepackage{diagbox}
\usepackage{float}
\usepackage{mathrsfs}

\def\BibTeX{{\rm B\kern-.05em{\sc i\kern-.025em b}\kern-.08em
    T\kern-.1667em\lower.7ex\hbox{E}\kern-.125emX}}

\algnewcommand{\algorithmicforeach}{\textbf{for each}}
\algdef{SE}[FOR]{ForEach}{EndForEach}[1]
  {\algorithmicforeach\ #1\ \algorithmicdo}
  {\algorithmicend\ \algorithmicforeach}

\DeclareMathOperator*{\argmin}{arg\,min}

\begin{document}

\title{Bi-capacity Choquet Integral for Sensor Fusion \\ with Label Uncertainty\\
\thanks{This material is based upon work supported by the National Science
Foundation under Grant IIS-2153171-CRII: III: Explainable Multi-Source Data Integration with Uncertainty.}
}

\author{\IEEEauthorblockN{Anonymous Authors}}

\author{\IEEEauthorblockN{Hersh Vakharia}
\IEEEauthorblockA{\textit{University of Michigan} \\
Ann Arbor, MI \\
hershv@umich.edu}
\and
\IEEEauthorblockN{Xiaoxiao Du}
\IEEEauthorblockA{\textit{University of Michigan} \\
Ann Arbor, MI \\
xiaodu@umich.edu}
}

\maketitle

\begin{abstract}
Sensor fusion  combines data from multiple sensor sources to improve reliability, robustness, and accuracy of data interpretation. The Fuzzy Integral (FI), in particular, the Choquet integral (ChI), is often used as a powerful nonlinear aggregator for fusion across multiple sensors. However, existing supervised ChI learning algorithms typically require precise training labels for each input data point, which can be difficult or impossible to obtain. Additionally, prior work on ChI fusion is often based only on the normalized fuzzy measures, which bounds the fuzzy measure values between $[0, 1]$. This can be limiting in cases where the underlying scales of input data sources are bipolar (i.e., between $[-1, 1]$). To address these challenges, this paper proposes a novel Choquet integral-based fusion framework, named \textit{Bi-MIChI} (pronounced ``bi-mi-kee''), which uses bi-capacities to represent the interactions between pairs of subsets of the input sensor sources on a bi-polar scale. This allows for extended non-linear interactions between the sensor sources and can lead to interesting fusion results. Bi-MIChI also addresses label uncertainty through Multiple Instance Learning, where training labels are applied to ``bags’’ (sets) of data instead of per-instance. Our proposed Bi-MIChI framework shows effective classification and detection performance on both synthetic and real-world  experiments for sensor fusion with label uncertainty. We also provide detailed analyses on the behavior of the fuzzy measures to demonstrate
our fusion process.
\end{abstract}


\section{Introduction}
Sensors are all around us collecting data. Each sensor may provide complementary or reinforcing information that supports tasks such as target detection, classification, or scene understanding. The fuzzy integral (FI), in particular, the Choquet integral (ChI), has been used widely as a powerful non-linear aggregator for sensor fusion \cite{grabisch1996application,gader2004multi,kakula2021online,skublewska2023aggregation, alvey2023geometrically}. The ChI is based on fuzzy measures, or ``capacities'', which are a set of real-valued parameters that represent the interactions between input sources and help determine the decision-making process.

Suppose we are fusing $m$ sensor data sources, $C=\{c_1,c_2,\cdots,c_m\}$. The power set of $C$ is denoted as $2^C$, which contains all possible crisp subsets of $C$. The Choquet integral fusion relies on a set of fuzzy measures (length $2^C$), where each fuzzy measure element, or ``capacity'', correspond to each subset of the sensor source combinations.  As an example, if the fuzzy measure is denoted as $\mathbf{g}$, the notation $\mathbf{g}_{\{1,2\}}$ indicates the the contributions from the intersection of sensor input source 1 and source 2. 

Two challenges exist for  Choquet integral-based fusion methods in the literature. First, most prior work on ChI fusion adopt the normalized fuzzy measures \cite{anderson2016fuzzy,du2016multiple,du2018multiple, narukawa2021score, zhang2021fuzzy}, which bounds the fuzzy measures between 0 and 1. To put it mathematically, the normalized fuzzy measure, $\mathbf{g}$, is a real-valued function that maps $2^C \rightarrow [0, 1]$.
This means, any combinations of the input sources are always weighted non-negatively. This can be troublesome as the normalized fuzzy measure may overlook  background information provided by the sensor sources. Here is an example to illustrate this effect. Consider a binary classification application where the ChI fusion is performed to detect humans/pedestrians in a scene. We want the human/pedestrian pixels to have target label ``1'' and everything else (background trees, buildings, and shadows, etc.) to have non-target label ``0''. Assume one of the input sensors is a shadow sensor that can detect shadows well (therefore, this sensor source has high values on shadow region and low values on ``human'' pixels). If the normalized fuzzy measure is used, it will assign value 0 to the measure element associated with the shadow sensor (in other words, ignore the shadow sensor input completely), as it did not contribute to the positive target (human) class. However, we argue that the shadow sensor also carries useful information that should not be ignored. If we know where the shadows are, we can apply a negative weight on the shadows to mark the ``non-human'' regions, and potentially improve the accuracy and remove shadow artifact from the pedestrian detection results.  Thus, it is important to explore extended fuzzy measures to accommodate and incorporate useful and complementary information from all sensor sources. 
In this work, we propose the use of bi-capacities \cite{grabisch2007bicapacitiespartI, grabisch2005bicapacitiespartII}, which are a generalization of fuzzy measures/capacities that are useful when data follows a bi-polar scale of $[-1,1]$. 

The bi-capacities naturally extend and generalize the ChI fusion to produce fusion results on a bipolar scale and can result in improved detection performance (as shown later in Experiments). 

Second, existing supervised ChI fusion algorithms often require precise instance-level labels for training. However, accurate labels can often be difficult, if not impossible, to obtain. Let us consider the pedestrian detection application again. It can be tedious and expensive to label each pixel on an image, but it is very simple to draw a bounding box region around where a person may appear. Based on the Multiple Instance Learning (MIL) framework \cite{dietterich1997solving}, we name such a region a ``bag'', where each bag contains a set of pixels. The MIL framework has been explored in previous literature for fuzzy fusion with bag-level label uncertainties \cite{ghosh2015fuzzy, du2018multiple}. However, none of these work handles fuzzy measures outside the $[0,1]$ range. This work is the first, to our knowledge, that addresses bi-capacity ChI fusion with bipolar sensor inputs and label uncertainty under the MIL framework.

The remainder of the paper is organized as follows. Section~\ref{sect:related} provides a review on related works and discusses prior work in ChI fusion and MIL formulation. Section~\ref{sect:bicap} presents the definition and properties of the bi-capacity ChI formulations. Section~\ref{sect:methodology} describes the proposed Bi-MIChI algorithm,  its objective functions,  and optimization strategies. Section~\ref{sect:experiements} presents experimental results of the proposed
algorithm on both simulated and real datasets for multi-sensor fusion. Section~\ref{sect:discussion}  discusses the main findings,  conclusions, and future work.

\section{Related Work} \label{sect:related}

\subsection{Bi-capacity \& Choquet Integral Applications}
Bi-capacities and bipolar capacities were developed in decision theory where the underlying scales are bipolar \cite{grabisch2007bicapacitiespartI}. They extend the concept of capacities, or fuzzy measures, by representing interactions of pairs of subsets on a bipolar scale, which can model multidimensional and complex problems. The bi-capacities and bipolar ChI has been explored previously in  \cite{grabisch2005bicapacitiespartII, abbas2014bipolar, abbas2021balancing}. However, these works mainly use hand-crafted, synthetic examples as an illustration and did not handle data and label uncertainties in real sensing data. To our knowledge, no prior experiments using bi-capacities have been conducted on real data integration applications considering imprecise labels. 

Bi-capacities have successfully been applied in multi-criteria decision making under uncertainty, where they can be used to aggregate positive and negative preferences \cite{ corrente2012promethee, zhang2017bitopsis, zhang2021microgrid}. Bi-capacities have also been used to perform complex, dynamic evaluations, such as evaluating regional eco-efficiency in \cite{peng2021regionaleco}. However, they have yet to be applied in the realm of sensor fusion as a way to represent the complex relationship between the input sources to be fused.

The Choquet Integral, on the other hand, has been used as an effective nonlinear aggregation operation for many applications, such as landmine detection \cite{gader2004multi, mendezvazquez2008minimum}, soil-type classification \cite{wang2015integration}, gesture recognition \cite{hirota2011multimodal},  and pedestrian detection \cite{guan2020classifier}. However, these applications do not utilize bi-capacities for modelling the complementary relationships between sources on a bipolar scale. This work will extend the ChI fusion to a bipolar scale using bi-capacities.

\subsection{Multiple Instance Learning}
The Multiple Instance Learning (MIL) framework was created to address the problem of label ambiguity and uncertainty in supervised learning \cite{dietterich1997solving}. Instead of requiring instance-level labels for supervised learning, MIL allows for uncertain labels for ``bags'' of instances. Bags are labeled positive if it contains at least one positive (``target'') instance and negative if all instances are negative (``non-target''). It has been successfully applied to a variety of applications, such as medical image and video analysis \cite{quellec2017medical}, landmine detection \cite{yuksel2015multiple, manandhar2015multiple, karem2011multiple}, scene segmentation \cite{kraus2016classify, cobb2017multiple}, and other remote sensing tasks \cite{zare2018multiple}. This work aims to extend MIL for bi-capacity ChI sensor fusion to handle label uncertainties, where the input labels are at bag-level, instead of instance-level (per pixel) labels.

\subsection{Multiple Instance Choquet Integral}
This work extends the work of Multiple Instance Choquet Integral (MICI) for classification and regression \cite{du2016multiple, du2018multiple}, which is a supervised ChI fusion framework under the MIL framework. One limitation of  MICI is that it utilizes normalized fuzzy measures bounded between $[0,1]$ and typically requires the input sensor data to be normalized between  $[0,1]$ as well. While MICI's normalized fuzzy measures can model some interactions between sources, the use of bi-capacities in this paper introduces new functionalities where some sensor source combinations can now be weighed negatively against others (which was not done before). This also leads to improved classification performance and enhanced interpretability. We also developed a new objective function to train the proposed Bi-MIChI algorithm and learn the bi-capacity values.

\section{Bi-capacity ChI Definitions}
\label{sect:bicap}
Bi-capacities are a variety of capacities, or fuzzy measures, that are useful when data follows a bipolar scale \cite{grabisch2007bicapacitiespartI}. When applied to fusion, they allow for interesting results where some sources, or combinations of sources, can be weighed negatively against others. Let $C = \{c_1, c_2, ..., c_m\}$ denote the $m$ sensor sources to be fused. A monotonic bi-capacity, $\mathbf{g}$, is a real valued function that maps $S(C) \rightarrow [-1, 1]$, where $S(C) = \{(A,B): A \subseteq C, B \subseteq C, A \cap B = \emptyset \}$.  $S(C)$ represents pairs of disjoint, crisp, subsets of $C$. It satisfies the following properties \cite{grabisch2005bicapacitiespartII}:
\begin{enumerate}
    \item Boundary Conditions: $$\mathbf{g}(\emptyset, \emptyset) = 0, \mathbf{g}(C, \emptyset) = 1, \mathbf{g}(\emptyset, C) = -1$$
    \item Monotonicity: $$\text{If} A \subseteq E \text{ and } B \supseteq F, \text{ then } \mathbf{g}(A, B) \geq \mathbf{g}(E, F)$$
\end{enumerate}

If we compare these properties to those of the normalized fuzzy measure (e.g., see Definition 2.1 of \cite{murofushi2000fuzzy}), we note the following specific properties for bi-capacities. First, the bi-capacities capture information of {\textit{ternary}} importance from multi-source data and have the potential of extracting useful and complementary information on bipolar scales. 
The sensor set $C$ can be configured into $3^m$ disjoint subset pairs, with $3^m-3$ non-boundary, learnable elements. Each element of the bi-capacity represents a weighting of the first subset ``against'' the second subset. This property allows for interesting fusion results, where some sources, or sets of sources, can be weighed negatively against other sources. Additionally, sources that have negative information can still be used to reinforce the final fusion result. In this paper, bi-capacity elements will be denoted with a subscript of its corresponding pair of subsets. For example, $\mathbf{g}_{1,23}$ corresponds to subset pair $(\{c_1\}, \{c_2, c_3\})$. In this example, the value of bi-capacity element $\mathbf{g}_{1,23}$ weighs $c_1$ {{against}} $c_2$ and $c_3$.

\begin{figure}[ht!]
    \centering
    \includegraphics[scale=0.28]{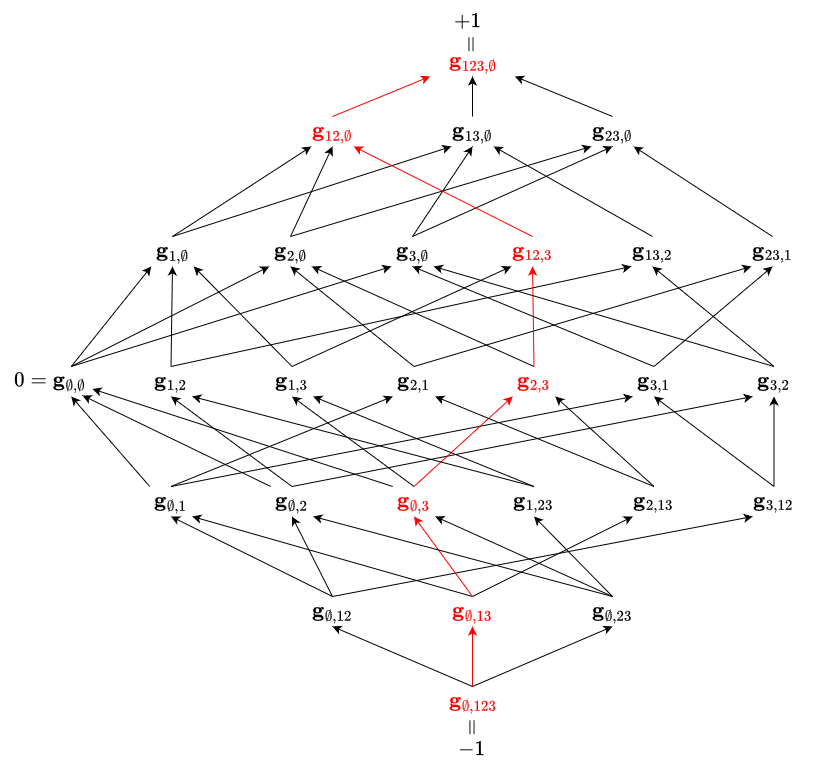}
    \caption{Illustration of bi-capacity element relationships given three sources. Red shows an example path of monotonicity.}
    \label{fig:bicap_flow}
\end{figure}

Second, to satisfy monotonicity, a bi-capacity \textit{increases} with inclusion in the first subset and \textit{decreases} with inclusion in the second subset. Fig. \ref{fig:bicap_flow} illustrates the monotonicity property of a bi-capacity for an example with three sources. As shown, bi-capacity elements $\mathbf{g}_{123,\emptyset}$ and $\mathbf{g}_{\emptyset,123}$ are the uppermost and lowermost bounds of $+1$ and $-1$, respectively. There are various paths between the bounds that satisfy monotonicity; one possible path is highlighted in red. As an example, take bi-capacity elements of $\mathbf{g}_{12,3}$ and $\mathbf{g}_{2,3}$, which correspond to subset pairs $(\{c_1,c_2\}, \{c_3\})$ and $(\{c_2\}, \{c_3\})$, respectively. The monotonicity property shows that $\mathbf{g}_{12,3}$ is an upper bound of $\mathbf{g}_{2,3}$, as $\{c_1,c_2\} \subseteq \{c_2\}$ and $\{c_3\} \supseteq \{c_3\}$.

Bi-capacities with three sources can also be represented using a three-dimensional lattice structure, as shown in Fig. 1 in \cite{grabisch2007bicapacitiespartI}. In this paper, the values of the bi-capacities are represented in matrix form, as shown in tables \ref{tab:um_bicap1} and \ref{tab:um_bicap2} in the experiments section. As seen in the tables, the bi-capacities create a repeating fractal pattern within the matrix \cite{grabisch2007bicapacitiespartI}.
\vspace{-4mm}

Given a bi-capacity, $\mathbf{g}$, instance-level fusion is computed with a using the discrete Choquet Integral, which is well-defined for bi-capacities \cite{grabisch2005bicapacitiespartII}. The bi-capacity Choquet Integral, $\mathscr{C}$, on instance $\mathbf{x}_n$ can be computed as
\begin{multline}
    \mathscr{C}_\mathbf{g}(\mathbf{x}_n) = \sum_{k=1}^{m} \biggl[ \bigl(|h(c_k; \mathbf{x}_n)| - |h(c_{k-1}; \mathbf{x}_n)|\bigr) \\
    \cdot \mathbf{g}(A_k \cap C^+, A_k \cap C^-) \biggr],
    \label{eq:ci}
\end{multline}
where $h(c_k; \mathbf{x}_n)$ denotes the output of the $k^{th}$ sensor source input, $c_k$, on the $n^{th}$ data instance, $\mathbf{x}_n$. $C$ is sorted such that $|h(c_1; \mathbf{x}_n)| \leq |h(c_2; \mathbf{x}_n)| \leq ... \leq |h(c_m; \mathbf{x}_n)|$. Since there is $m$ classifiers, $h(c_{0}; \mathbf{x}_n)$ is defined to be 0. When using the bi-capacity $\mathbf{g}$ with the Choquet Integral, it is defined that $A_k = \{c_k, ..., c_m\}$, where $C$ is sorted. Additionally, $C^+ = \{i \in C \mid h(c_i; \mathbf{x}_n) \geq 0\}$ and $C^- = C \setminus C^+$.

\section{The Proposed Bi-MIChI} \label{sect:methodology}
To utilize the Choquet Integral successfully for fusion, the non-boundary elements of the bi-capacity must be learned from a set of training data. In the MIL framework, data is organized in ``bags'' with uncertain and imprecise labels. Based on the MIL assumptions, bags are labeled positive if at least one instance in the bag is positive and negative if all instances in that bag are negative \cite{dietterich1997solving}. 

Including or excluding the $\mathbf{g}_{\emptyset, \emptyset} = 0$ bound can provide different results for the bi-capacity ChI. If this bound is removed, elements of $\mathbf{g}_{\emptyset, \cdot}$ can be positive and $\mathbf{g}_{\cdot, \emptyset}$ can be negative, which allows for an inversion effect on the data. In the following, we present two variations on the objective functions, one without and one with the $\mathbf{g}_{\emptyset, \emptyset}$ term, and describe our optimization strategies to learn the bi-capacities from the training data and bag-level labels.

\subsection{Objective Function 1: without $\mathbf{g}_{\emptyset, \emptyset}=0$}
\label{sect:obj1}
The objective function for the negative and positive bags can be written as

\begin{equation}
    \begin{aligned}
        J^- = \sum_{a=1}^{B^-} \max_{\forall \mathbf{x}_{ai}^- \in \mathcal{B}_a^-} \left(\mathscr{C}_\mathbf{g}(\mathbf{x}_{ai}^-) + 1 \right)^2,
    \end{aligned}
    \label{eq:j_minus}
\end{equation}
\begin{equation}
    \begin{aligned}
        J^+ = \sum_{b=1}^{B^+} \min_{\forall \mathbf{x}_{bj}^+ \in \mathcal{B}_b^+} \left(\mathscr{C}_\mathbf{g}(\mathbf{x}_{bj}^+) -1\right)^2,
    \end{aligned}
    \label{eq:j_plus}
\end{equation}
where $B^+$ is the number of positive bags, $B^-$ is the number of negative bags, $\mathbf{x}_{ai}^-$ is the $i^{th}$ instance in the $a^{th}$ negative bag, $\mathcal{B}_a^-$, and $\mathbf{x}_{bj}^+$ is the $j^{th}$ instance in the $b^{th}$ positive bag, $\mathcal{B}_b^+$. $\mathscr{C}_\mathbf{g}$  is the Choquet Integral output for a bi-capacity, $\mathbf{g}$. 

This objective function is parallel to the min-max model for normalized fuzzy measures \cite{du2018multiple}. However, the bi-capacities adds additional ternary information (as defined in Section~\ref{sect:bicap}) and the computation of bi-capacity ChI is different compared with prior work and now has the range between $[-1, 1]$. The bi-capacity, $\mathbf{g}$, is obtained by minimizing the objective function in equation \ref{eq:objective}. The $J^-$ term  in Eq.\eqref{eq:j_minus} encourages the Choquet integral of instances in the negative bags to be ``$-1$'', while the $J^+$ term in Eq.\eqref{eq:j_plus} encourages the instances in the positive bags to ``$+1$''. Therefore, given training data with positively and negatively labeled bags, minimizing $J$ results in a bi-capacity that will produce a high confidence on the target and low confidence on the background. The overall objective can be written as
\begin{equation}
    \begin{aligned}
        \min_\mathbf{g} J = J^- + J^+ .
    \end{aligned}
    \label{eq:objective}
\end{equation}

\subsection{Objective Function 2: with $\mathbf{g}_{\emptyset, \emptyset}=0$}
\label{sect:obj2}
Since the $\mathbf{g}_{\emptyset, \emptyset}=0$ bound enforces $\mathbf{g}_{\cdot, \emptyset} \geq 0$ and $\mathbf{g}_{\emptyset, \cdot} \leq 0$, the ChI is unable to produce an inversion effect on negative data. Therefore, a different objective function must be used to properly represent the negative data in the fused result.
We can define a new set of $J^+$  and $J^-$ as 
\begin{equation}
    \begin{aligned}
        J^- = \sum_{a=1}^{B^-} \max_{\forall \mathbf{x}_{ai}^- \in \mathcal{B}_a^-} \left(\mathscr{C}_\mathbf{g}(\mathbf{x}_{ai}^-) - 0 \right)^2,
    \end{aligned}
    \label{eq:j_minus_new}
\end{equation}
\begin{equation}
    \begin{aligned}
        J^+ = \sum_{b=1}^{B^+} \min_{\forall \mathbf{x}_{bj}^+ \in \mathcal{B}_b^+} \left(1 - \left |\mathscr{C}_\mathbf{g}(\mathbf{x}_{bj}^+)\right| \right)^2.
    \end{aligned}
    \label{eq:j_plus_new}
\end{equation}

Instead of having $J^-$ encourage negative bags to have a ChI output of $-1$, we instead encourage them to the neutral value of 0. Additionally, $J^+$ now encourages data in the positive bags to be close to the poles of $-1$ and $+1$. The idea is that the sources to be fused may detect the target class negatively or positively. Thus, instead of trying to invert the negative detection, we can represent the target as values that are on either pole.
Then, after fusion, the absolute value of the result can be taken to reflect high confidence in the target. The overall objective is the sum of these two terms, as defined in Eq.~\eqref{eq:objective}.

\subsection{Optimization Algorithm}

\begin{figure*}[t]
    \centering
    \includegraphics[width=0.9\linewidth]{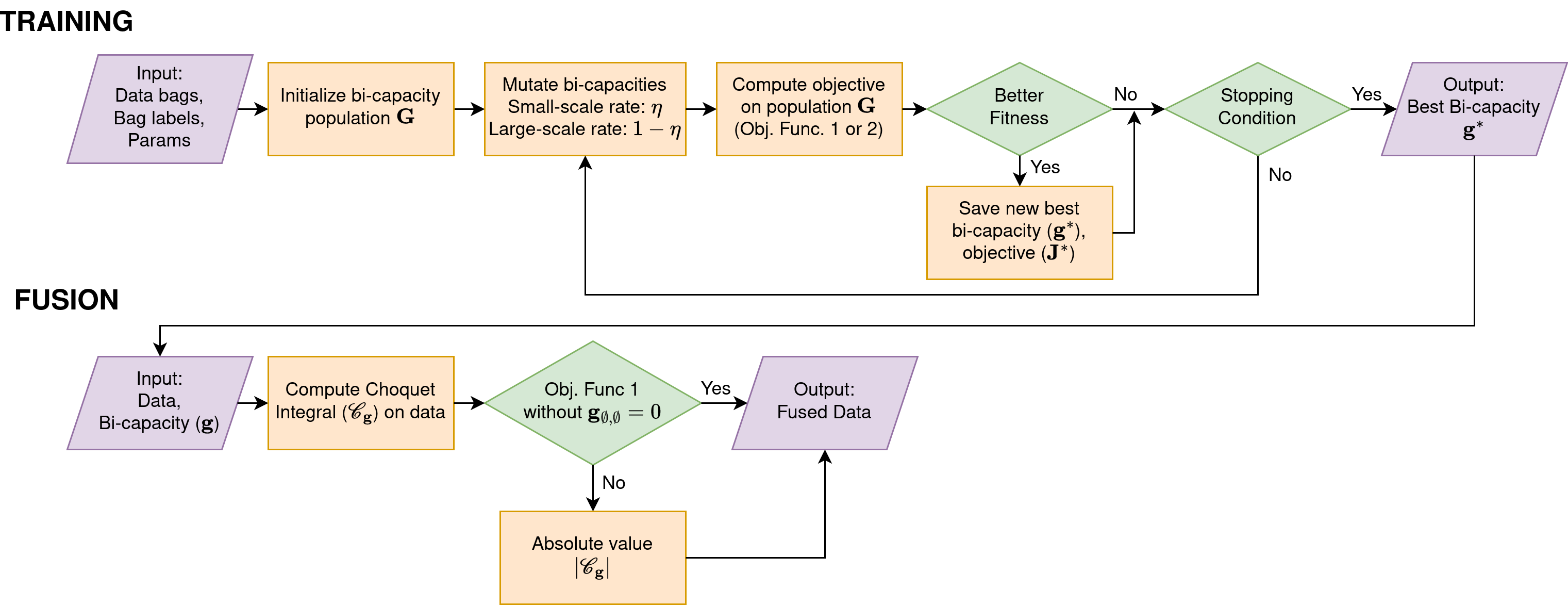}
    \caption{Flowchart for the Bi-MIChI  algorithm.}
    \label{fig:alg_flowchart}
\end{figure*}

The optimization is achieved through the use of an evolutionary algorithm following that of MICI \cite{du2018multiple}, with several modifications based on bi-capacities. Algorithm \ref{alg:bicap-mici} shows the pseudo-code of the optimization and fusion steps for the proposed Bi-MIChI algorithm, and Table~\ref{tab:algo_legend} explains the parameters used. Fig. \ref{fig:alg_flowchart} shows a flowchart of the algorithm. The algorithm requires training data (sensor inputs from multiple sources) and bag-level labels for the training bags as inputs. The output of the algorithm is the learned bi-capacities, which will be used to generate instance-level fusion results  given multi-sensor sources.

First, a population (size $P$) of bi-capacities is initialized. The experiments in this paper used $P=36$. Bi-capacities are initialized starting at the $\mathbf{g}_{\emptyset, \emptyset} = 0$ boundary and moving outward toward the upper and lower bounds. This process is completed similarly to breadth-first-search exploration \cite{cormen2022introduction}, where the neighbor nodes are the upper and lower bounds of the current node. As an example, when $\mathbf{g}_{\emptyset, \emptyset}$ is included in Objective Function 2, after starting at $\mathbf{g}_{\emptyset, \emptyset}$, the lower bound elements $\mathbf{g}_{1, \emptyset}$, $\mathbf{g}_{2, \emptyset}$, and $\mathbf{g}_{3, \emptyset}$ and upper bound elements $\mathbf{g}_{\emptyset, 1}$, $\mathbf{g}_{\emptyset, 2}$, and $\mathbf{g}_{\emptyset, 3}$ are the ``neighbors'' that are to be explored and sampled next. The current element is sampled uniformly randomly between the minimum of all the previously sampled upper bounds and the maximum of all the previously sampled lower bounds. This same methodology is used for if the $\mathbf{g}_{\emptyset, \emptyset}=0$ boundary is not used (Objective Function 1); the 0 bound is simply not enforced during sampling.

In the main optimization loop, bi-capacities are updated using a combination of small-scale and large-scale mutations. In the small-scale mutation, only one of the bi-capacity elements is re-sampled. The element to be sampled is determined through a multinomial distribution based one the number of times each element of the bi-capacity is used by the Choquet Integral during training. Therefore, the element that is used the most during training will have the highest probability to be resampled. Eq.~\eqref{eq:multinomial} shows the formula for the probability for resampling, where $v_{A,B}$ is the number of times bi-capacity element $\mathbf{g}_{A,B}$ is used during training. The selected element is resampled uniformly between the associated upper and lower bounds. In the large-scale mutation, an entirely new bi-capacity is sampled. The probability of small and large mutations is determined by a user-defined parameter $\eta$. Small mutations occur with a probability of $\eta$, and large mutations occur with a probability of $1-\eta$. The value $\eta=0.8$ was found to be effective empirically during experimentation.

\begin{equation}
\begin{aligned}
P(\mathbf{g}_{A,B}) = \frac{v_{A, B}}{\sum_{\mathbf{g}_{E,F} \in \mathbf{g}} v_{E,F}}
\end{aligned}
\label{eq:multinomial}
\end{equation}

After mutation, the fitness of each bi-capacity is computed using the objective function~\eqref{eq:objective}. Bi-capacities with improved fitness are saved. This process of mutation and fitness calculation continues for a maximum of $I$ iterations ($I=5000$ in this paper's experiments) or until certain stopping criteria is met. In our experiments, the stopping criteria is set to be if the new best fitness is within a threshold, $\mathbf{J}_T$, of the previous best fitness (in other words, if the fitness does not change any more than the $\mathbf{J}_T$ threshold). We chose a small threshold value $\mathbf{J}_T = 0.001$ as it yielded effective convergence results in our experiments.  

\begin{algorithm}[t]
\caption{Bi-MIChI Algorithm}
\label{alg:bicap-mici}
\begin{algorithmic}[1] 
\Statex \textbf{TRAINING}
\Require Training Data, Training Labels, Parameters 
\State Initialize a population of bi-capacities: $\mathbf{G}$
\State $\mathbf{J}^* = \min(\mathbf{J}^0_P)$, $\mathbf{g}^* = \argmin_\mathbf{G}(\mathbf{J}^0_P)$
\For{ $i:=1 \to I$ }   
		\For{ $p:=1 \to P$ }
			\State Randomly generate $z \in [0, 1]$.
			\If {$z < \eta $}  
				\State Small-scale mutation of $\mathbf{G}_p$
			\Else
				\State Large-scale mutation of $\mathbf{G}_p$  
			\EndIf
		\EndFor
	
    \If {using $\mathbf{g}_{\emptyset, \emptyset} = 0$}
        \State Compute $\mathbf{G}$ fitness using \eqref{eq:j_minus_new} and \eqref{eq:j_plus_new} \algorithmiccomment{\textcolor{blue}{Object Function 2}}
    \Else
        \State Compute $\mathbf{G}$ fitness using  \eqref{eq:j_minus} and \eqref{eq:j_plus}
        \algorithmiccomment{\textcolor{blue}{Object Function 1}}
    \EndIf
 
	\State Compute $ \mathbf{J}_d = |\min(\mathbf{J}_P^{i}) - \mathbf{J}^*|$.
	\If {$ \min(\mathbf{F}_P^{i}) < \mathbf{J}^*$}
					\State $\mathbf{J}^* = \max(\mathbf{J}_P^{i})$, $\mathbf{g}^* = \argmin_\mathbf{G}(\mathbf{J}^i_P) $.
	\EndIf
	\If {$\mathbf{J}_d \leq \mathbf{J}_T$}
	break;
	\EndIf
\EndFor

\Return $\mathbf{g}^* $

\Statex \textbf{RUN FUSION}
\Require Data, Bi-capacity ($\mathbf{g}$)
\If {using $\mathbf{g}_{\emptyset, \emptyset} = 0$} \algorithmiccomment{\textcolor{blue}{Object Function 2}}
    \State \Return $|\mathscr{C}_\mathbf{g}(\text{Data})|$
\Else  \algorithmiccomment{\textcolor{blue}{Object Function 1}}
    \State \Return $\mathscr{C}_\mathbf{g}(\text{Data})$
\EndIf

\end{algorithmic}
\end{algorithm}

\begin{table}[h!]
    \centering
      \caption{Algorithm \ref{alg:bicap-mici} list of parameters.}
    \label{tab:algo_legend}
    \resizebox{\hsize}{!}{
    \begin{tabular}{|c|c|}
        \hline
        Parameter & Explanation \\
        \hline
        $\mathbf{G}$ & Population of bi-capacities\\
        $P$ & Bi-capacity population size\\
        $\mathbf{J}^*$ & Best fitness value \\
        $\mathbf{g}^*$ & Best bi-capacity \\
        $\mathbf{J}^i_P$ & Fitnesses of population of bi-capacities at iteration $i$ \\
        $\mathbf{J}_T$ & Fitness threshold for stopping condition
        $i$\\
        $I$ & Maximum optimization iterations\\
        $\eta$ & Small-scale mutation probability\\
         \hline
    \end{tabular}
    }
\end{table}

\section{Experiments} \label{sect:experiements}
The proposed Bi-MIChI algorithm is applied to a synthetic and a real-world
classification task. Experimental results are presented to illustrate the fusion performance of the proposed algorithm. 

\subsection{Synthetic Dataset}
The first experiment showcases the fusion abilities on a synthetic dataset. This dataset was constructed to highlight the proposed framework’s ability to utilize negative information using bi-capacities, as well as its ability to handle bag-level imprecise labels. Two sets of tests are run on this dataset, the first without the $\mathbf{g}_{\emptyset, \emptyset}=0$ bound using objective functions \eqref{eq:j_minus} and \eqref{eq:j_plus} (Section~\ref{sect:obj1}), and the second uses the $\mathbf{g}_{\emptyset, \emptyset}=0$ bound and objective functions \eqref{eq:j_minus_new} and $\eqref{eq:j_plus_new}$ (Section~\ref{sect:obj2}).

To generate this simulated example, suppose the data contains a letter ``U'' and a letter ``M'' in the scene. We chose these two letter shapes as they are symmetric and easy to visualize. The ground truth is shown in Fig.~\ref{fig:um_gt}. Assume we are fusing three sensor inputs to detect both shapes. The first sensor source only detects the letter shape ``U'' (as shown in Fig.~\ref{fig:um_source1}). The second sensor source highlights all background except the letter shape ``M'' (Fig.~\ref{fig:um_source2}). The third sensor source detects all background (non-letter) pixels (Fig.~\ref{fig:um_source3}). This example is inspired by real-world target detection applications, where the ``U'' and ``M''-shaped targets  consist of different materials, say, the ``U''-shape is metal and ``M''-shape is plastic, and the goal is to detect both targets. The source 1 detector may only be able to detect the metal target shape ``U'', whereas sensor 2 and 3 can only detect vegetation and background (non-target) pixels in the scene. 

To generate bag-level data, the SLIC superpixel algorithm \cite{achanta2012slic} is used to group the pixels in the image into bags (superpixels). The generated bags are shown in Fig. \ref{fig:um_bags}. Bags that contains any part of the ``U'' or ``M'' targets are labeled positive (target class, highlighted as green), and the remaining bags are labeled as negative (non-target class, shown as red). 

\subsubsection{Simulated Experiment 1 with Bipolar Labels}
The first test case is shown in Fig. \ref{fig:um_sources_results1}. The sources for fusion are shown in Fig. \ref{fig:um_source1}-\ref{fig:um_source3}. The goal is to learn a bi-capacity that is able to detect both ``U'' and ``M'' shapes with high confidence. In this experiment, bipolar-scale labels are used, where the target class (``U'' and ``M'') are pushed to label $+1$ and the non-target class (background) is pushed to label $-1$, as shown in the ground truth figure (Fig.~\ref{fig:um_gt}).

Fig.~\ref{fig:um_fused} shows the proposed Bi-MIChI fusion result using Objective Function 1 (bipolar case). We observed some interesting behaviors for the bi-capacities. As shown in Fig.\ref{fig:um_fused}, the ``U'' shape was detected perfectly (has a positive fusion label of +1), but the ``M'' shape in the final fusion result was forced to the value $-1$. This is because the ``M'' shape in all three sources are not detected, i.e., the input sensor source values take the form $[-1, -1, -1]$. In this case, the Choquet integral pushes all negatives $[-1, -1, -1]$ to the $-1$ boundary, which resulted in a missed detection for the ``M'' letter. 

Table~\ref{tab:um_bicap1} shows the learned bi-capacity values. The bi-capacity is notated using the form $\mathbf{g}_{A,B}$, which represents a weighting of the sensor source subset  A “against” the second subset B (see the bi-capacity definition in Section~\ref{sect:bicap}). The bold elements mark the measure element values that were actually used. Note that not all measure elements were updated based on how the sensor input data is sorted in the ChI computation (see Eq.~\ref{eq:ci}).  We observed that, in this experiment, the bi-capacity element $\mathbf{g}_{23, 1} = -0.85$, which represents the combination of sources 2 \& 3 both \textit{negatively} detect some parts of the target, while source 1 positively detects the target. The bi-capacity measure element $\mathbf{g}_{23, 1}$ has a high magnitude, which shows that the negative contribution of the intersection between sources 2 \& 3 \textit{against} the positive contribution of source 1 are weighted highly to help detect the target shapes. This makes sense, as the combination of source 1, the negative of source 2, and the negative of source 3 all contribute to detect both ``U'' and ``M'' shapes.

\begin{figure}[t!]
    \centering
    \includegraphics[width=0.9\columnwidth]{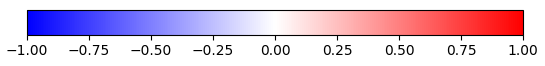}
    \caption{Colorbar for experimental results.}
    \label{fig:colorbar}
\end{figure}

\begin{figure}[h]
  \centering
  \begin{subfigure}[b]{0.3\columnwidth}
    \includegraphics[width=\linewidth]{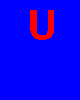}
    \caption{Source 1}
    \label{fig:um_source1}
  \end{subfigure}
  \hfill
  \begin{subfigure}[b]{0.3\columnwidth}
    \includegraphics[width=\linewidth]{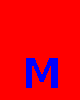}
    \caption{Source 2}
    \label{fig:um_source2}
  \end{subfigure}
  \hfill
  \begin{subfigure}[b]{0.3\columnwidth}
    \includegraphics[width=\linewidth]{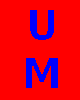}
    \caption{Source 3}
    \label{fig:um_source3}
  \end{subfigure}

\medskip 

  \begin{subfigure}[b]{0.3\columnwidth}
    \includegraphics[width=\linewidth]{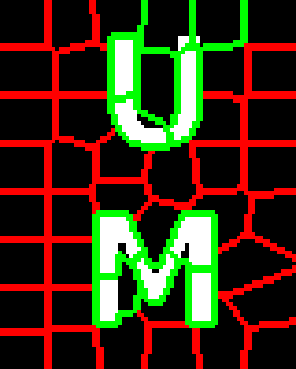}
    \caption{Labeled Bags}
    \label{fig:um_bags}
  \end{subfigure}
  \hfill
  \begin{subfigure}[b]{0.3\columnwidth}
    \includegraphics[width=\linewidth]{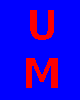}
    \caption{Ground Truth}
    \label{fig:um_gt}
  \end{subfigure}
  \hfill
  \begin{subfigure}[b]{0.3\columnwidth}
    \includegraphics[width=\linewidth]{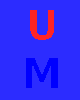}
    \caption{Fusion 1}
    \label{fig:um_fused}
  \end{subfigure}

\medskip
    \begin{subfigure}[b]{0.3\columnwidth}
    \includegraphics[width=\linewidth]{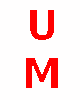}
    \caption{Ground Truth 2}
    \label{fig:um_gt_new}
  \end{subfigure}
  \hfill
  \begin{subfigure}[b]{0.3\columnwidth}
    \includegraphics[width=\linewidth]{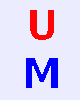}
    \caption{\scriptsize{Fusion 2 w/o Abs}}
    \label{fig:um_out_new}
  \end{subfigure}
  \hfill
  \begin{subfigure}[b]{0.3\columnwidth}
    \includegraphics[width=\linewidth]{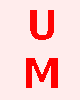}
    \caption{Fusion 2}
    \label{fig:um_out_abs_new}
  \end{subfigure}

  \caption{The Synthetic  Experiment results. (a)-(c) are the sources for fusion, (d) shows positive bags in green and negative bags in red, (e) shows the ground truth on the bipolar scale, and (e) shows the fusion result of our Bi-MIChI using objective function 1. (g) shows the ground truth where the non-negative is neutral (label 0). (h) shows the fusion results of our Bi-MIChI using objective function 2, but without the absolute value on the Bi-capacity Choquet term in Eq.~\eqref{eq:j_plus_new}. (i) shows the fusion results of our Bi-MIChI using the complete objective function 2. All images follow the colorbar in Fig. \ref{fig:colorbar}.}
  \label{fig:um_sources_results1}
\end{figure}

\begin{table}[h]
    \centering
   \caption{The learned bi-capacity with Simulated Experiment 1 (bipolar labels). Bold elements are updated during ChI fusion.}
    \label{tab:um_bicap1}
    \resizebox{\hsize}{!}{
    \begin{tabular}{|c|cccccccc|}
     \hline
     \multicolumn{9}{|c|}{UM Experiment 1 Bi-capacity, $\mathbf{g}_{A,B}$} \\
     \hline
     \diagbox{A}{B}  & $\emptyset$ & 1 & 2 & 12 & 3 & 13 & 23 & 123\\
     \hline
     $\emptyset$ & 0 & -0.96 & -0.94 & -1.00 & \textbf{-0.87} & -0.98 & \textbf{-1.00} & \textbf{-1.00} \\
     1 & 0.45 &  & -0.91 &  & -0.10 &  & -0.95 & \\
     2 & 0.55 & -0.94 &  &  & \textbf{0.39} & -0.96 &  & \\
     12 & 1.00 &  &  &  & \textbf{0.97} &  &  & \\
     3 & \textbf{0.10} & -0.86 & -0.82 & -0.88 &  &  &  & \\
     13 & 0.73 &  & -0.72 &  &  &  &  & \\
     23 & \textbf{0.77} & \textbf{-0.85} &  &  &  &  &  & \\
     123 & 1.00 &  &  &  &  &  &  & \\
     \hline
\end{tabular}}
\end{table}

\subsubsection{Simulated Experiment 2 with $[0,1]$ Labels}
The second test case is shown in Fig.~\ref{fig:um_gt_new}, \ref{fig:um_out_new}, and \ref{fig:um_out_abs_new}. We implement the Object Function 2 with $\mathbf{g}_{\emptyset, \emptyset}=0$ and ran Bi-MIChI  for fusion.  The input sources and bags are the same as the previous experiment. In this test case, however, the ground truth non-target bags are now pushed to a neutral value of 0 (instead of -1 in the previous experiment). The target region (the ``U'' and ``M'') are now pushed to either $-1$ or $+1$ (in other words, the absolute value of the target class is pushed to $1$). The resulting ground truth map is shown in Fig.~\ref{fig:um_gt_new}, where the ``U'' and ``M'' shapes are still highlighted. The ChI fusion result is shown in Fig. \ref{fig:um_out_new}, and the absolute value result is shown in Fig. \ref{fig:um_out_abs_new}. As seen in the figure, the fused result is able to detect the ``UM'' with high confidence, showing that pushing positive bags to the $-1$ and $+1$ poles can be more successful.

Table~\ref{tab:um_bicap2} shows the learned bi-capacity values for UM experiment 2. In contrast to experiment 1, we observe that the bi-capacity element $\mathbf{g}_{23, 1} = 0.07$. The magnitude of this element is close to 0, which has the effect of zeroing out the negative contribution of the intersection of sources 2 \& 3 against the positive contribution of source 1. This makes sense, as it contributes to the effect of pushing background information to 0. Furthermore, as previously stated, the ``M'' data of the three sources takes the form $[-1, -1, -1]$, which the ChI pushes to $-1$. However, taking the absolute value allows for a high a confidence detection on the ``M'' as well.

\begin{table}[h]
\centering
\caption{The learned bi-capacities with Simulated Experiment 2 ($[0,1]$ labels).}
\label{tab:um_bicap2}
\resizebox{\hsize}{!}{
\begin{tabular}{|c|cccccccc|}
     \hline
     \multicolumn{9}{|c|}{UM Experiment 2 Bi-capacity, $\mathbf{g}_{A,B}$} \\
     \hline
     \diagbox{A}{B}  & $\emptyset$ & 1 & 2 & 12 & 3 & 13 & 23 & 123\\
     \hline
     $\emptyset$ & 0 & -0.15 & -0.09 & -0.33 & \textbf{-0.62} & -0.89 & \textbf{-0.97} & \textbf{-1.00} \\
     1 & 0.92 &  & 0.81 &  & 0.90 &  & 0.80 & \\
     2 & 0.83 & 0.03 &  &  & \textbf{0.31} & 0.02 &  & \\
     12 & 0.99 &  &  &  & \textbf{0.99} &  &  & \\
     3 & \textbf{0.91} & -0.04 & 0.84 & -0.04 &  &  &  & \\
     13 & 0.95 &  & 0.84 &  &  &  &  & \\
     23 & \textbf{0.96} & \textbf{0.07} &  &  &  &  &  & \\
     123 & 1.00 &  &  &  &  &  &  & \\
     \hline
\end{tabular}
}
\end{table}

\subsection{Low-Light Pedestrian Detection}

\begin{figure}[h]
  \centering
  \includegraphics[width=\columnwidth]{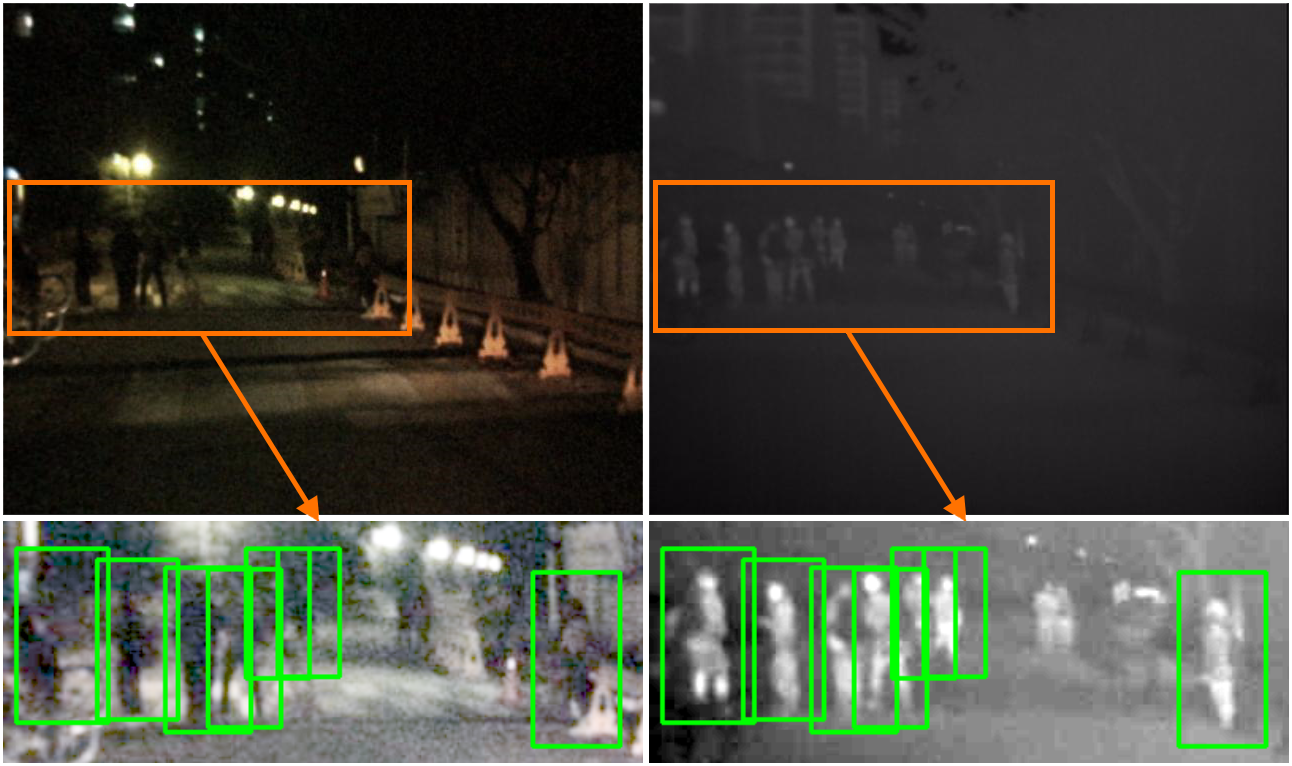}
  \caption{KAIST Pedestrian RGB and thermal image pair with ground truth bounding boxes. The images containing the green boxes are over exposed for easier viewing.}
  \label{fig:ped_rgb_therm_bbox}
\end{figure}

We conduct additional experiments on a real-world classification task using the KAIST Multispectral Pedestrian Detection Benchmark \cite{hwang2015multispectral}. The KAIST dataset contains RGB and thermal sensors and the goal is to detect pedestrians in the scene. We selected a low-light setting to illustrate the necesscity for sensor fusion, as the thermal camera will be able to highlight pedestrians in low-light whereas the RGB camera will contribute negatively to the detection (RGB cameras will show dark backgrounds with some spots from light sources such as street lamps). Fig.~\ref{fig:ped_rgb_therm_bbox} shows an illustration of an image from the KAIST dataset. As shown, under the low light conditions, pedestrians are very difficult to detect, and it is necessary to fuse information from both RGB and thermal cameras. Additionally, since pedestrians are very difficult to see, it is almost impossible to annotate and produce precise pixel-level training labels. However, it is possible to generate bounding boxes (as shown in green in Fig.~\ref{fig:ped_rgb_therm_bbox}) to indicate  possible pedestrian locations. In this experiment, we use the green bounding boxes as the bag-level labels to train our Bi-MIChI algorithm for RGB and thermal fusion. 

The low-light conditions also motivate the use of the bi-capacity's negative weighting. While the thermal image is able to produce high confidence on the pedestrians, there are false positive readings on the background buildings as well as lights. Conversely, the RGB image shows low confidence on the pedestrians, but a high confidence on well-lit areas, like the lights, foreground, and street cones. Negatively weighing the RGB image can contribute to a high confidence on the pedestrians, without relying too heavily on the thermal data.

\begin{figure}
  \centering
  \begin{subfigure}[b]{0.48\columnwidth}
    \centering
    \includegraphics[width=\linewidth]{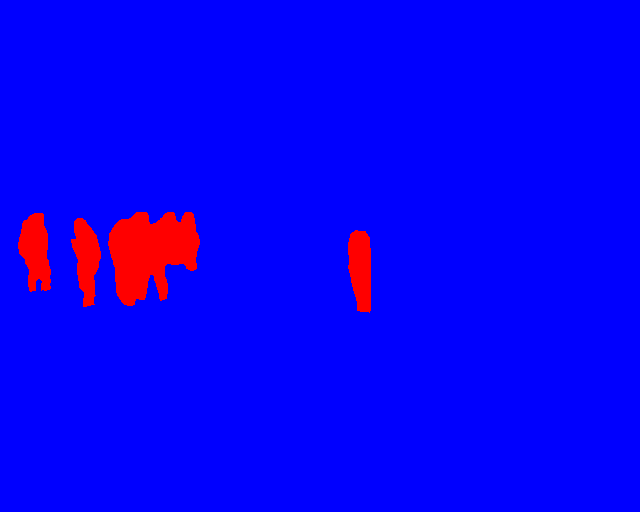}
    \caption{Pedestrian Ground Truth}
    \label{fig:ped_mask_gt}
  \end{subfigure}
  \hfill
  \begin{subfigure}[b]{0.48\columnwidth}
    \centering
    \includegraphics[width=\linewidth]{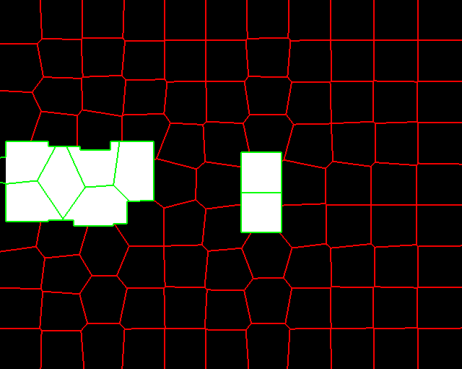}
    \caption{Labeled bags}
    \label{fig:ped_bags}
  \end{subfigure}

  \caption{Pedestrian ground truth and labeled bags. Positive bags are shown in green, and negative bags are shown in red. The bags were generated from the bounding box masks.}
  \label{fig:ped_gt_bags}
\end{figure}

\begin{figure}[t]
  \centering
  \begin{subfigure}[b]{0.32\columnwidth}
    \includegraphics[width=\linewidth]{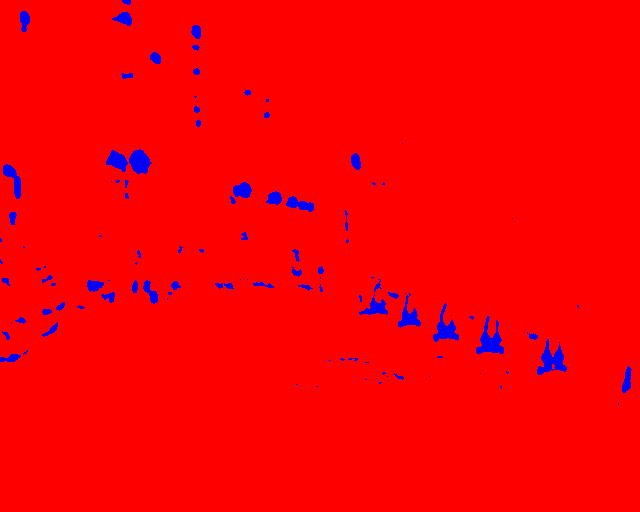}
    \caption{Source 1}
    \label{fig:ped_source1}
  \end{subfigure}
  \hfill
  \begin{subfigure}[b]{0.32\columnwidth}
    \includegraphics[width=\linewidth]{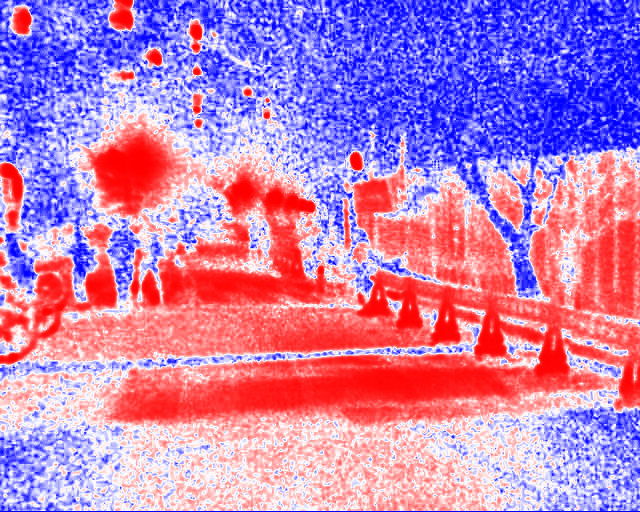}
    \caption{Source 2}
    \label{fig:ped_source2}
  \end{subfigure}
  \hfill
  \begin{subfigure}[b]{0.32\columnwidth}
    \includegraphics[width=\linewidth]{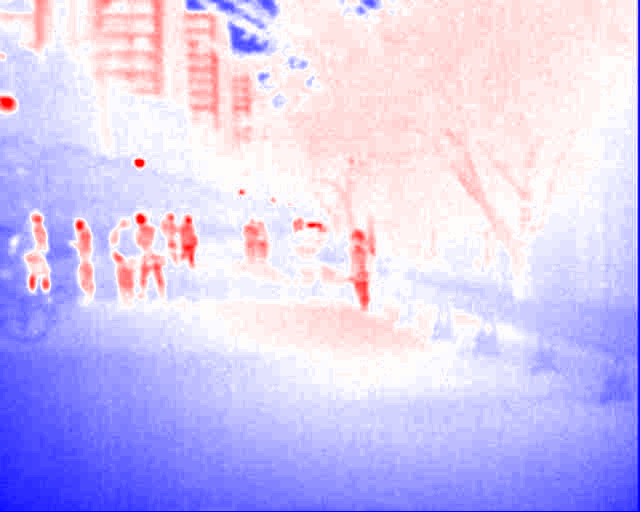}
    \caption{Source 3}
    \label{fig:ped_source3}
  \end{subfigure}

\medskip 

  \begin{subfigure}[b]{0.32\columnwidth}
    \includegraphics[width=\linewidth]{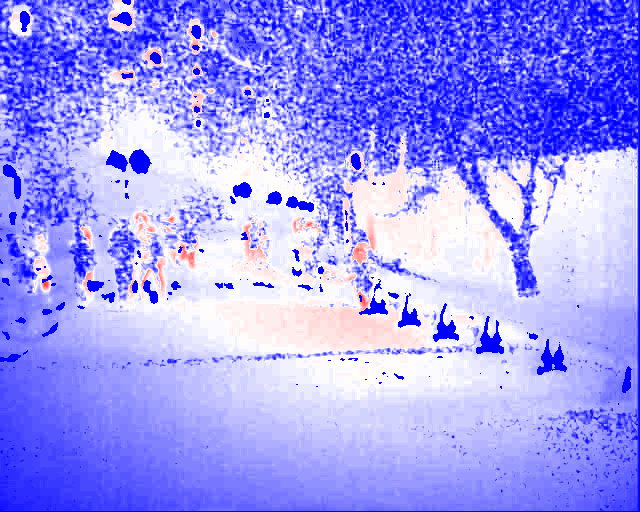}
    \caption{Min}
    \label{fig:ped_min}
  \end{subfigure}
  \hfill
  \begin{subfigure}[b]{0.32\columnwidth}
    \includegraphics[width=\linewidth]{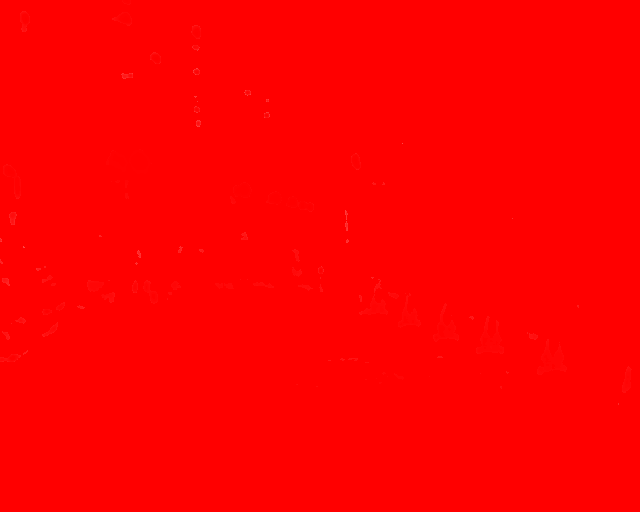}
    \caption{Max}
    \label{fig:ped_max}
  \end{subfigure}
  \hfill
  \begin{subfigure}[b]{0.32\columnwidth}
    \includegraphics[width=\linewidth]{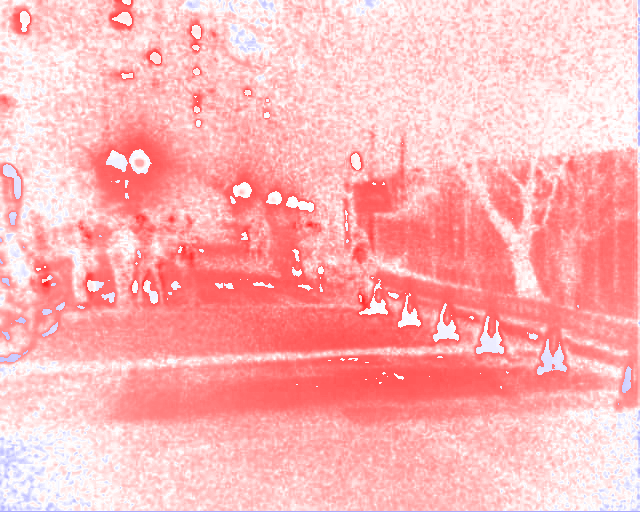}
    \caption{Mean}
    \label{fig:ped_mean}
  \end{subfigure}   

\medskip

  \begin{subfigure}[b]{0.32\columnwidth}
    \includegraphics[width=\linewidth]{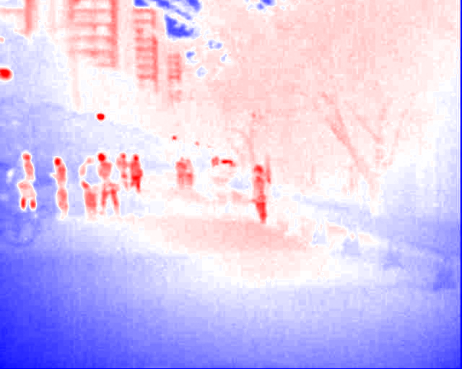}
    \caption{Weighted Mean}
    \label{fig:ped_wmean}
  \end{subfigure}
  \hfill
  \begin{subfigure}[b]{0.32\columnwidth}
    \includegraphics[width=\linewidth]{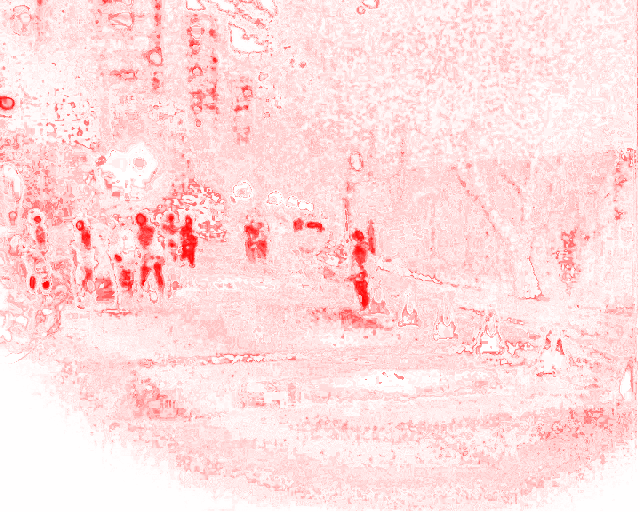}
    \caption{KNN}
    \label{fig:ped_knn}
  \end{subfigure}
  \hfill
  \begin{subfigure}[b]{0.32\columnwidth}
    \includegraphics[width=\linewidth]{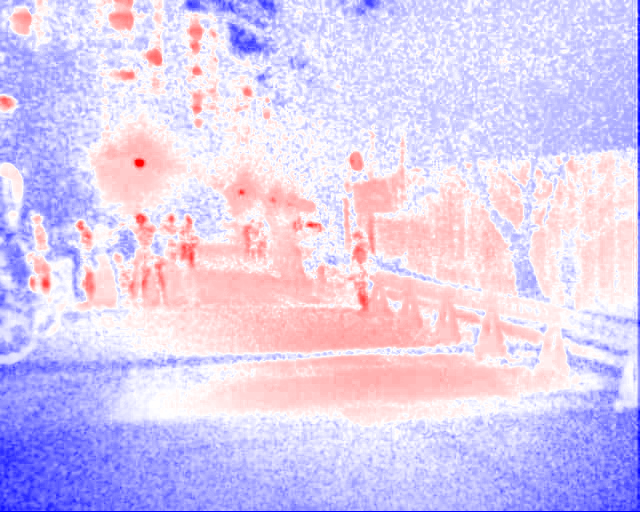}
    \caption{SVM}
    \label{fig:ped_svm}
  \end{subfigure}
  
\medskip

  \begin{subfigure}[b]{0.32\columnwidth}
    \includegraphics[width=\linewidth]{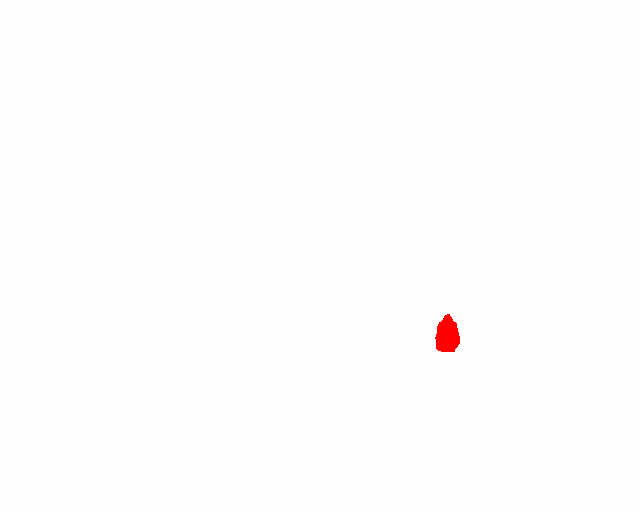}
    \caption{Mask RCNN}
    \label{fig:ped_maskrcnn}
  \end{subfigure}
  \hfill
  \begin{subfigure}[b]{0.32\columnwidth}
    \includegraphics[width=\linewidth]{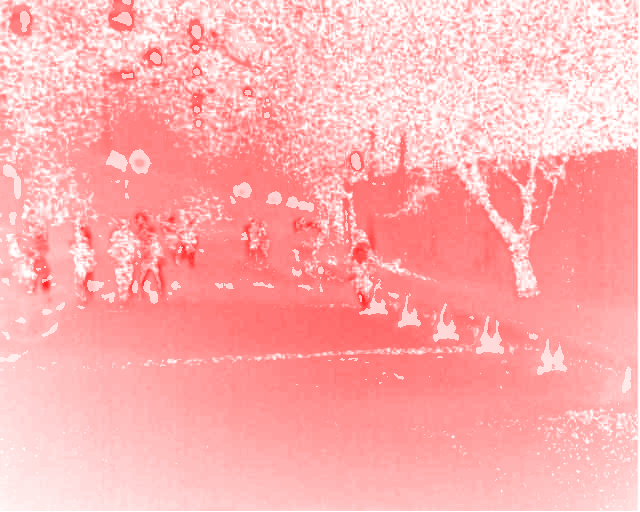}
    \caption{CI-QP}
    \label{fig:ped_ciqp}
  \end{subfigure}
  \hfill
  \begin{subfigure}[b]{0.32\columnwidth}
    \includegraphics[width=\linewidth]{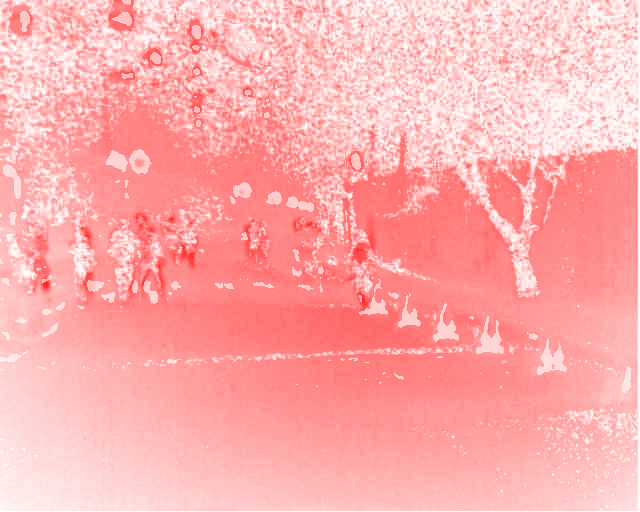}
    \caption{MICI}
    \label{fig:ped_noisyor}
  \end{subfigure}

\medskip

  \begin{subfigure}[b]{0.32\columnwidth}
    \includegraphics[width=\linewidth]{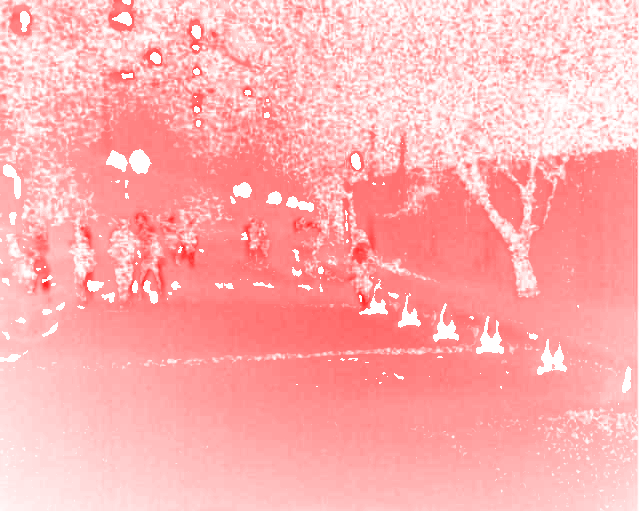}
    \caption{MICI-BFM}
    \label{fig:ped_bfm}
  \end{subfigure}
  \hfill
  \begin{subfigure}[b]{0.32\columnwidth}
    \includegraphics[width=\linewidth]{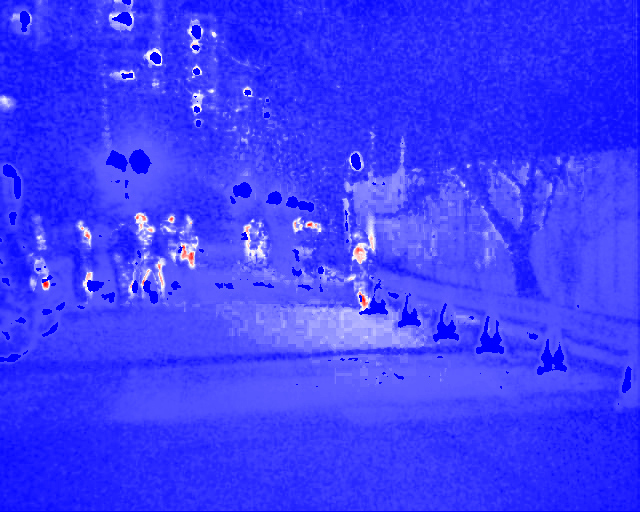}
    \caption{Bi-MIChI (Obj. 1)}
    \label{fig:ped_bicap_obj1}
  \end{subfigure} 
  \hfill
  \begin{subfigure}[b]{0.32\columnwidth}
    \includegraphics[width=\linewidth]{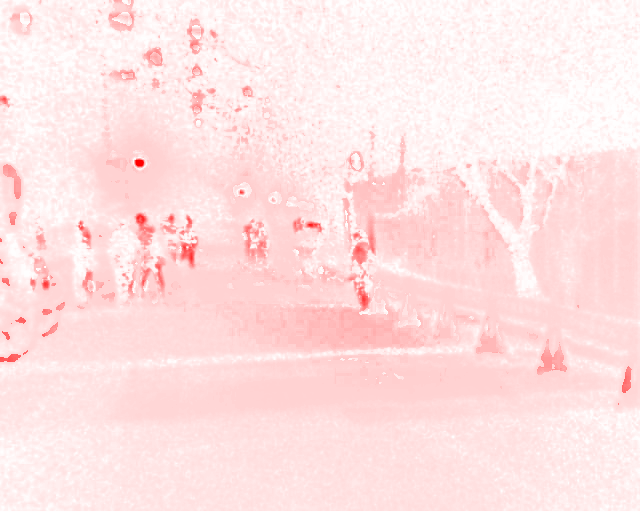}
    \caption{Bi-MIChI (Obj. 2)}
    \label{fig:ped_bicap_obj2}
  \end{subfigure}  
  \caption{Pedestrian detection fusion results. All results follow the colorbar in Fig. \ref{fig:colorbar}. Best viewed in color.}
  \label{fig:ped_sources_results}
\end{figure}

The SLIC algorithm \cite{achanta2012slic} is used to generate superpixels (``bags'') for training. Since the dataset provides ground truth in the form of bounding boxes, bags that contain points that intersect with any of the bounding boxes are considered positive, as shown in Fig. \ref{fig:ped_bags}. Fig. \ref{fig:ped_mask_gt} shows the pixel-level ground truth labels. Note that the pixel-level ground truth labels were not used during training, and were only used to evaluate our final fusion result.

Fig. \ref{fig:ped_sources_results} shows the three sensor sources that are used as fusion inputs, as well as comparison results for fusion. Source 1 (Fig.~\ref{fig:ped_source1}) utilizes thresholding on a grayscaled version of the RGB image to negatively detect bright areas the image on the visual spectrum. Source 2 (Fig.~\ref{fig:ped_source2}) is a grayscaled version of the RGB image with histogram equalization. This shows low, negative confidence on pedestrians and tree, and high confidence on the lights, traffic cones, shadows, and other road elements. Source 3 (\ref{fig:ped_source3}) comes from the thermal data with gamma correction. This source shows high confidence on pedestrians, but there are a few false positives in the background that must be reduced.

The proposed Bi-MIChI was used to perform fusion on this dataset using both Objective Function 1 and Objective Function 2. We compare the fusion results to a variety of comparison methods. First, the min, max, and mean is taken across three sources as a simple fusion baseline. A weighted mean was calculated using weights determined by minimizing the mean squared error between the fusion result and the ground truth. Then, K-nearest neighbors (KNN) with $k=100$ and Support Vector Machine (SVM) approaches were used to perform pixel-level fusion. We also compare it to a pre-trained Mask R-CNN \cite{he2018mask}, which is a neural network-based object detector. Mask R-CNN was run on the low-light RGB image and the segmentation mask is of detected elements of the ``person'' class. While Mask R-CNN has seen great success in computer vision applications, pre-trained models can exhibit poor performance in low-light applications where object details are unclear without significant image processing. We also ran comparisons with other  Choquet Integral-based methods. CI-QP \cite{grabisch1996application} learns a fuzzy measure for fusion by formulating the problem as a quadratic program with instance-level labels. MICI is a previous MIL method for fusion  but uses normalized fuzzy measures with elements values in the $[0,1]$ range. MICI-BFM \cite{du2019multiple} is a version of the MICI that uses binary fuzzy measures, which are fuzzy measure with elements that only take values $\{0,1\}$. KNN, SVM, and CI-QP all require instance level labels, while MICI, MICI-BFM, and Bi-MIChI operate on bag-level labels. 

Visual results for all methods are shown in Fig. \ref{fig:ped_sources_results}. It is important to note  that all results are the direct output of the fusion algorithm - some output a confidence map on the $[-1,1]$ scale, while others output on a $[0,1]$ scale.  Naively taking the min, max, and mean of the sources produces poor qualitative results, as the sources highlight conflicting information (background vs. person). The weighted mean produced a result that is very similar to source 3 - this is because the resulting weights heavily favored the third source. KNN  highlighted the pedestrians fairly well, but there is a considerable amount of background noise. The KNN method is also highly sensitive to the parameter choice of $k$. When we reduced the k value (e.g., when $k=5$), the fusion performance deteriorated significantly and the output became much noisier. Similarly, the SVM approach was able to highlight some pedestrians, but also showed some false positives that occur in the source images. Mask R-CNN falsely identifies a traffic cone as a pedestrian, showing its limitations in low-light conditions. CI-QP, MICI, and MICI-BFM all produce similar results, all of which captured background information with fairly high confidence. Qualitatively, we can see that the proposed Bi-MIChI with objective function 1 is effective at eliminating background information, while still detecting the pedestrians correctly. Bi-MIChI with objective function  2 produces similar results, but contains more false positives compared to objective function 1.

The Bi-MIChI bi-capacities learned with objective function 1 and 2 are shown in Tables \ref{tab:ped_bicap} and \ref{tab:ped_bicap2}, respectively. We can observe that many of elements in the bi-capacity learned with objective function 1 are set to $-1$. This occurs to encourage the background elements to have an output of $-1$. The positive detections are the result of the upper boundary element, $\mathbf{g}_{123, \emptyset}$. For the objective function 2 bi-capacities in Table \ref{tab:ped_bicap2}, we can observe some elements that help push the background to 0 (Recall that Objective 2 pushes the non-target class to a neutral label of 0). For example, elements $\mathbf{g}_{1, \emptyset}$, $\mathbf{g}_{2, \emptyset}$, and $\mathbf{g}_{12, \emptyset}$, which are the weights in favor of sources 1 and 2, are closer to 0. This makes sense, as those sources show high confidence in background elements. This is in contrast to the previous bi-capacities, in which there is an inversion effect.

The normalized fuzzy measures learned by CI-QP, MICI, and MICI-BFM are shown in Table \ref{tab:ped_fuzzy_measures}. The measures learned by these methods are similar as they all rely primarily on the element $\mathbf{g}_{123}$, while the other measure element values are close to 0. This suggests that, unlike bi-capacities, normal fuzzy measures are unable to properly take advantage of the negative or background information in the sources, particularly from sources 1 and 2 in this experiment.

\begin{table}[t!]
\centering
\caption{Pedestrian detection bi-capacity learned with objective function 1. Bolded elements are used in the Bi-MIChI.}
\label{tab:ped_bicap}
\resizebox{\hsize}{!}{
\begin{tabular}{|c|cccccccc|}
     \hline
     \multicolumn{9}{|c|}{Pedestrian Detection Bi-capacity (Obj. 1), $\mathbf{g}_{A,B}$} \\
     \hline
     \diagbox{A}{B}  & $\emptyset$ & 1 & 2 & 12 & 3 & 13 & 23 & 123\\
     \hline
     $\emptyset$ & 0 & \textbf{-1.00} & \textbf{-1.00} & -1.00 & \textbf{-1.00} & -1.00 & \textbf{-1.00} & -1.00 \\
     1 & \textbf{-0.82} &  & \textbf{-0.95} &  & \textbf{-0.92} &  & \textbf{-0.98} & \\
     2 & \textbf{-0.99} & \textbf{-1.00} &  &  & -0.99 & \textbf{-1.00} &  & \\
     12 & \textbf{-0.70} &  &  &  & \textbf{-0.79} &  &  & \\
     3 & \textbf{-1.00} & -1.00 & \textbf{-1.00} & -1.00 &  &  &  & \\
     13 & \textbf{-0.21} &  & \textbf{-0.59} &  &  &  &  & \\
     23 & \textbf{-0.99} & \textbf{-1.00} &  &  &  &  &  & \\
     123 & \textbf{1.00} &  &  &  &  &  &  & \\
     \hline
\end{tabular}
}
\end{table}

\begin{table}[ht!]
\centering
\caption{Pedestrian detection bi-capacity learned with objective function 2. Bolded elements are used in the Bi-MIChI.}
\resizebox{\hsize}{!}{
\begin{tabular}{|c|cccccccc|}
     \hline
     \multicolumn{9}{|c|}{Pedestrian Detection Bi-capacity (Obj. 2), $\mathbf{g}_{A,B}$} \\
     \hline
     \diagbox{A}{B}  & $\emptyset$ & 1 & 2 & 12 & 3 & 13 & 23 & 123\\
     \hline
     $\emptyset$ & 0 & \textbf{-0.26} & \textbf{-0.40} & -0.59 & \textbf{-0.68} & -1.00 & \textbf{-0.69} & -1.00 \\
     1 & \textbf{0.11} &  & \textbf{0.00} &  & \textbf{0.09} &  & \textbf{0.00} & \\
     2 & \textbf{0.07} & \textbf{-0.09} &  &  & -0.57 & \textbf{-0.98} &  & \\
     12 & \textbf{0.19} &  &  &  & \textbf{0.15} &  &  & \\
     3 & \textbf{0.63} & 0.61 & \textbf{0.02} & -0.30 &  &  &  & \\
     13 & \textbf{0.79} &  & \textbf{0.02} &  &  &  &  & \\
     23 & \textbf{1.00} & \textbf{1.00} &  &  &  &  &  & \\
     123 & \textbf{1.00} &  &  &  &  &  &  & \\
     \hline
\end{tabular}
}
\label{tab:ped_bicap2}
\end{table}

\begin{table}[h]
\centering
\caption{Measures learned by comparison Choquet integral methods (MICI, MICI-BFM, CI-QP).}
\label{tab:ped_fuzzy_measures}
\resizebox{\hsize}{!}{
\begin{tabular}{|c|ccccccc|}
     \hline
     Method & $\mathbf{g}_1$ & $\mathbf{g}_2$ & $\mathbf{g}_3$ & $\mathbf{g}_{12}$ & $\mathbf{g}_{13}$ & $\mathbf{g}_{23}$ & $\mathbf{g}_{123}$\\
     \hline
     MICI & 0.00 & 0.00 & 0.04 & 0.00 & 0.04 & 0.40 & 1\\
     MICI-BFM & 0 & 0 & 0 & 0 & 0 & 0 & 1\\
     CI-QP & 0 & 0 & 0 & 0 & 0 & 0.36 & 1\\
     \hline
\end{tabular}
}
\end{table}

\begin{table}[]
    \centering
    \caption{The AUC and RMSE results for the pedestrian detection fusion experiment. \textbf{Best}, \underline{Second Best} across ChI and fuzzy measure-based methods.}
  \label{tab:ped_results}
    \resizebox{\hsize}{!}{
    \begin{tabular}{c|c|c}
         \hline
         \textbf{Fusion Method} & 
         \textbf{AUC }$\uparrow$ & \textbf{RMSE} $\downarrow$\\
         \hline
         RGB Adaptive Thresholding (Source 1) & 0.495 & 1.948 \\
         RGB Histogram Equalization (Source 2) & 0.470 & 1.168 \\
         Thermal Gamma Correction (Source 3) & {0.812} & 0.897  \\
         \cdashline{1-3} 
         Min & 0.634 & 0.717  \\
         Max & 0.494 & 1.966  \\
         Mean & 0.621 & 1.279 \\
         Weighted Mean & 0.813 & 0.928 \\
         KNN & {0.795} & 0.405 \\
         SVM & 0.740 & 0.936 \\
         Mask R-CNN & 0.499 & {0.375} \\
         \cdashline{1-3} CI-QP & 0.640 & 0.806 \\
         MICI Noisy-or & 0.650 & 0.810 \\
         MICI-BFM & 0.634 & 0.805 \\
        \textbf{Bi-MIChI} (Obj. 1) & \textbf{0.711} & \textbf{0.388} \\
\textbf{Bi-MIChI} (Obj. 2) & \underline{0.654} & \underline{0.390}\\
         \hline
    \end{tabular}
}
\end{table}

To quantitatively compare the different fusion methods, the Receiver Operating Characteristic (ROC) curve's Area Under Curve (AUC) and Root Mean Square Error (RMSE) metrics are computed between the result and the ground truth shown in Fig.\ref{fig:ped_mask_gt}. A higher AUC signifies better target detection, while a lower RMSE signifies that the result is closer to the ground truth. Mask R-CNN, CI-QP, MICI, MICI-BFM, and our Bi-MIChI with objective function 2 produce confidence maps that are bounded from $[0,1]$. These methods were normalized from $[-1,1]$ before the metrics were computed to offer a consistent comparison between all methods.

Table \ref{tab:ped_results} shows the results of these metrics. Interestingly, source 3 from the thermal sensor after gamma correction, along with weighted mean (which closely matches source 3), obtained a very high AUC for pedestrian detection. This is because the thermal sensor is able to sense the heat signature from humans even under low-light conditions. However, all input sources (including source 3) have a high RMSE value, which means all of them contain a lot of noise and false positives, especially in the non-pedestrian areas. Among the fusion methods, KNN and SVM  performed really well on AUC, but both of these methods require pixel-level training labels and are highly sensitive to parameter choices. The Mask R-CNN method produced very low RMSE, but based on the visual result (Fig.~\ref{fig:ped_maskrcnn}), Mask R-CNN hardly detected any pedestrians and simply classified almost everything to background (thus, a low AUC). Among the Choquet integral-based methods (bottom five rows),  our proposed Bi-MIChI achieves superior performance with higher AUC and lower RMSE compared to other Choquet integral-based methods using normalized fuzzy measures. This also shows that the bi-capacities proposed in this work are effective for detecting target classes (humans) while removing background noise, given input sensor sources that carry complementary information. Additionally, our proposed Bi-MIChI are trained with only bag-level labels, which allows the fusion process to take label uncertainty into consideration.

\subsection{Discussions} 

This paper tests two objective functions with the proposed Bi-MIChI algorithm. The first does not enforce the $g_{\emptyset, \emptyset} = 0$ bound when constructing the bi-capacities, while the second does. In our experiments, we find that they are both useful for different applications. Objective function 2 works well if separate sources highlight different parts of the target. For example, in the synthetic ``U''``M'' experiment that uses objective function 2, the ``U'' is detected strongly positively in source 1 and the ``M'' is detected strongly negatively in source 2. Instead of trying to push both to $1$ like objective function 1 would do, each letter can be pushed to the closest pole (+1 for ``U'' and -1 for ``M'').

On the other hand, objective function 1 can be useful in situations where multiple sources detect the target, but on different ends of the $[-1,1]$ scale. For example, in the pedestrian detection experiment, pedestrians appear negatively in source 2 and positively in source 3. The ``inversion'' created by the objective function 2 allows the background to be pushed to $-1$ while still encouraging the pedestrians to label $+1$.

The use of bi-capacities and the Choquet Integral offers an explainable representation of the interactions between the sources to be fused. By examining the element values of the bi-capacities, users can interpret what role a sensor source plays in the final fusion result. Other machine learning techniques such as KNN, SVM, and Mask R-CNN, while effective in some cases, could be considered a ``black box'' in the sense that they provide predictions without a clear, interpretable reason a decision was made. The Choquet Integral, on the other hand, provides a clear, analytical model for complex, nonlinear fusion, and the bi-capacities provide  an interpretable and explainable representation of how fusion sources are weighed against each other.

\section{Conclusion}
\label{sect:discussion}

This paper presents Bi-MIChI, a novel Choquet integral-based method for explainable and interpretable data fusion with uncertain labels using bi-capacities. Through both synthetic and real-world experiments, this framework showed promising results in its ability to utilize negative and complementary information from input sensor sources for effective multi-sensor fusion.

Bi-MIChI aims to provide a general framework of data fusion that can be applied to various modalities and applications. To further explore the effectiveness of this method, additional experiments can be conducted applying Bi-MIChI to other fusion applications and sensor modalities. For example, this method can be used for multi-temporal and multi-view fusion \cite{alvey2023geometrically}, where a sequence of images can be fused for target detection and classification. 

\bibliographystyle{IEEEbib}
\small
\bibliography{references.bib}

\end{document}